\documentclass[lettersize,journal]{IEEEtran}
\usepackage{amsmath,amsfonts,amssymb}
\usepackage{array}
\usepackage{textcomp}
\usepackage{stfloats}
\usepackage{url}
\usepackage{verbatim}
\usepackage{graphicx}
\usepackage{caption}  
\usepackage{subcaption} 
\usepackage{cite}
\usepackage{comment}
\usepackage{float}
\usepackage{xcolor}
\usepackage{multirow}
\usepackage{booktabs}
\usepackage{algorithm}
\usepackage{algpseudocode}
\usepackage{graphicx}
\usepackage{subcaption}

\begin{document}

\title{A Multisource Fusion Framework for Cryptocurrency Price Movement Prediction}

\author{Saeed Mohammadi Dashtaki, Reza Mohammadi Dashtaki, Mehdi Hosseini Chagahi, Behzad Moshiri (Senior Member, IEEE), Md. Jalil Piran (Senior Member, IEEE)

\thanks{(Corresponding Authors: B. Moshiri and M. J. Piran)}

\thanks{S. M. Dashtaki and M. H. Chagahi are with the School of Electrical and Computer Engineering, College of Engineering, University of Tehran, Tehran, Iran, (e-mail: saeedmohammadi.d@ut.ac.ir; mhdi.hoseini@ut.ac.ir)}

\thanks{R. M. Dashtaki is with the Department of Chemistry, Isfahan University of Technology, Isfahan, Iran, (e-mail: sr.mohammadi@ch.iut.ac.ir)}

\thanks{B. Moshiri is with the School of Electrical and Computer Engineering, College of Engineering, University of Tehran, Tehran, Iran and the Department of Electrical and Computer Engineering, University of Waterloo,
Waterloo, Canada, (e-mail: moshiri@ut.ac.ir; bmoshiri@uwaterloo.ca)}

\thanks{M. J. Piran is with the Department of Computer Science and Engineering, Sejong University, Seoul 05006, South Korea, (e-mail: piran@sejong.ac.kr)}
}




\maketitle

\begin{abstract}
Predicting cryptocurrency price trends remains a major challenge due to the volatility and complexity of digital asset markets. Artificial intelligence (AI) has emerged as a powerful tool to address this problem. This study proposes a multisource fusion framework that integrates quantitative financial indicators, such as historical prices and technical indicators, with qualitative sentiment signals derived from X (formerly Twitter). Sentiment analysis is performed using Financial Bidirectional Encoder Representations from Transformers (FinBERT), a domain-specific BERT-based model optimized for financial text, while sequential dependencies are captured through a Bidirectional Long Short-Term Memory (BiLSTM) network. Experimental results on a large-scale Bitcoin dataset demonstrate that the proposed approach substantially outperforms single-source models, achieving an accuracy of approximately 96.8\%. The findings underscore the importance of incorporating real-time social sentiment alongside traditional indicators, thereby enhancing predictive accuracy and supporting more informed investment decisions.
\end{abstract}


\section{Introduction}
\label{introduction}
\IEEEPARstart{C}{ryptocurrencies} have rapidly emerged as influential players in global financial markets, attracting significant attention from both investors and researchers. Their growth has been remarkable, with the total market capitalization reaching nearly two trillion dollars in early 2022 \cite{xiang2023babd, zhou2023multi}. Since the introduction of Bitcoin in 2008, the ecosystem has evolved into an open-source community and decentralized payment network \cite{nakamoto2008bitcoin, shah2023vital}, followed by the emergence of a wide range of digital currencies. As this market continues to expand at an accelerated pace, the demand for effective forecasting methods to guide investment decisions has increased substantially. 

Recent advances in artificial intelligence (AI) and machine learning (ML) have shown strong potential for modeling complex, nonlinear financial systems \cite{ma2023multi, chagahi2024explainable, ashtiani2023news, htay2025enhancing}. In particular, deep learning (DL) architectures such as Long Short-Term Memory (LSTM) and Gated Recurrent Units (GRU) have been widely employed to capture sequential dependencies in financial time series \cite{nazareth2023financial, sahu2023overview, majidi2024algorithmic, dashtaki2022stock}. At the same time, transformer-based language models have become essential for analyzing unstructured text, enabling sentiment analysis of financial news and social media \cite{critien2022bitcoin, hosain2025hybrid}. Despite these advances, predictive performance remains constrained when models rely solely on a single type of data source. 

Historical prices and technical indicators provide objective and systematic signals for trend forecasting. At the same time, investor sentiment and market discourse from platforms such as X (formerly Twitter) significantly influence short-term cryptocurrency dynamics \cite{aslam2022sentiment, critien2022bitcoin}. Although inherently noisy, social data provide valuable insights into collective market behavior and complement traditional financial indicators.  

Data fusion techniques provide a promising avenue for integrating heterogeneous information streams. While data fusion combines raw signals, information fusion emphasizes the integration of structured, interpreted features to achieve greater reliability and precision \cite{meng2020survey, becerra2021information}.

This study proposes a multisource information fusion framework that integrates both quantitative technical indicators and qualitative sentiment signals for cryptocurrency price movement prediction. Sentiment features are extracted using Financial Bidirectional Encoder Representations from Transformers (FinBERT), a financial domain–specific variant of BERT, while sequential dependencies are modeled using a Bidirectional Long Short-Term Memory (BiLSTM) network. By jointly leveraging historical price patterns and real-time market sentiment, the proposed approach provides a more comprehensive understanding of cryptocurrency dynamics.

The key contributions of this work are as follows:
\begin{itemize}
\item A unified multisource framework is introduced that fuses technical indicators with sentiment signals derived from financial news and tweets, enabling more robust cryptocurrency trend prediction.

\item The proposed approach is evaluated on a comprehensive large-scale Bitcoin dataset spanning more than seven years, which integrates historical market data with sentiment information collected from reputable Bitcoin-related news accounts on X.

\item Extensive experiments demonstrate that the proposed BiLSTM–FinBERT model significantly outperforms state-of-the-art baselines, achieving 96.8\% accuracy and 99.6\% AUC, while also yielding superior financial returns in backtesting.

\item Our findings highlight the practical value of incorporating social sentiment into trading strategies, offering enhanced predictive accuracy and improved profitability for investors and algorithmic systems.
\end{itemize}

The remainder of the paper is organized as follows. Section \ref{relatedwork} reviews related studies on cryptocurrency forecasting. Section \ref{Material and methods} introduces the proposed framework. Section \ref{Experiments and Results} presents experimental setups, results, and discussion. Section \ref{Future Research Directions} outlines potential extensions, and Section \ref{Conclusion} concludes the study.

\section{Related Work}
\label{relatedwork}

A wide range of approaches has been explored to forecast cryptocurrency prices, with varying degrees of success. The accuracy of prediction models is largely determined by the quality and relevance of the input data, making the selection of informative features a critical factor.

Early studies primarily focused on quantitative market indicators, including daily price information and technical indicators such as the moving average (MA), relative strength index (RSI), and moving average convergence divergence (MACD) \cite{ortu2022technical, pabucccu2023forecasting}. These indicators capture underlying trading behaviors and price patterns that have historically been used to guide forecasting models.

More recent research has incorporated unstructured textual data, particularly financial news and social media content, alongside numerical indicators. By applying natural language processing (NLP) techniques, these studies aimed to extract market sentiment and behavioral signals to enhance predictive performance \cite{khan2024crypto, roumeliotis2024llms}. For example, Li et al. \cite{li2023market} integrated quantitative metrics with sentiment-derived news features into ML classifiers for predicting 10-day price movements. Azamjon et al. \cite{azamjon2023forecasting} proposed a Q-learning algorithm that combined on-chain data with whale-alert tweets to forecast Bitcoin trends. Similarly, Zhong et al. \cite{zhong2023lstm} combined historical trading patterns with sentiment analysis from social media, employing an LSTM network with a relationwise graph attention mechanism (ReGAT).

Building on this line of work, Htay et al. \cite{htay2025enhancing} proposed several LSTM-based variants that integrate social media sentiment and trading volume with historical prices, showing that sentiment-enhanced models significantly outperform traditional univariate LSTM approaches in Bitcoin forecasting. Similarly, Hosain et al. \cite{hosain2025hybrid} introduced xFiTRNN, a hybrid architecture that integrates FinBERT embeddings with BiGRU and self-attention, enhanced by linearized phrase structures, to improve both predictive performance and explainability in financial sentiment analysis.

Although single-source models provide useful insights, they often fail to capture the multifaceted dynamics of cryptocurrency markets. Multisource approaches that integrate both quantitative and qualitative signals generally achieve improved robustness and predictive accuracy. Nevertheless, existing multisource methods have frequently been limited by narrow data coverage, suboptimal feature design, or insufficient modeling of temporal dependencies. Addressing these challenges requires more comprehensive frameworks capable of jointly leveraging diverse sources of information. To this end, the next section introduces a multisource fusion framework that integrates technical indicators with sentiment signals extracted from financial news and social media.

\section{The Proposed Model}
\label{Material and methods}

This section introduces the proposed model, which combines heterogeneous information sources to improve the accuracy of cryptocurrency price movement predictions. The framework integrates both quantitative technical indicators and qualitative sentiment features derived from financial news and social media, employing advanced ML techniques to capture intricate patterns and trends within the data.

\subsection{Overview of the Model}
The model is designed to predict cryptocurrency price movements by incorporating two primary data components: (i) quantitative information consisting of historical price data and technical indicators, and (ii) qualitative information extracted from social media platforms such as X. These two sources are fused to improve prediction performance by leveraging the strengths of both numerical and textual features.

A BiLSTM architecture is employed, chosen for its ability to capture long-term dependencies in sequential data by processing information in both forward and backward directions. This bidirectional approach allows the model to grasp the full temporal context, enabling the detection of patterns and trends that might be missed by unidirectional models.

Price movement prediction is formulated as a binary classification task, indicating an ``up'' or ``down'' movement on the following trading day. The target label can be defined as:

\begin{equation}
    y_{t} =
    \begin{cases}
    1 &  \quad C_{t} > C_{t-1}, \\
    0 & \quad \text{otherwise,}
    \end{cases} \label{eq:trend}   
\end{equation}

where $y_{t}$ denotes the cryptocurrency price movement, with ``1'' representing an upward trend and ``0'' a downward trend. $C_{t}$ is the closing price of the cryptocurrency on day $t$, and $C_{t-1}$ is the closing price on day $t-1$ \cite{zhong2023lstm}.

\begin{figure*}[t]
    \centerline{\includegraphics[scale=0.40]{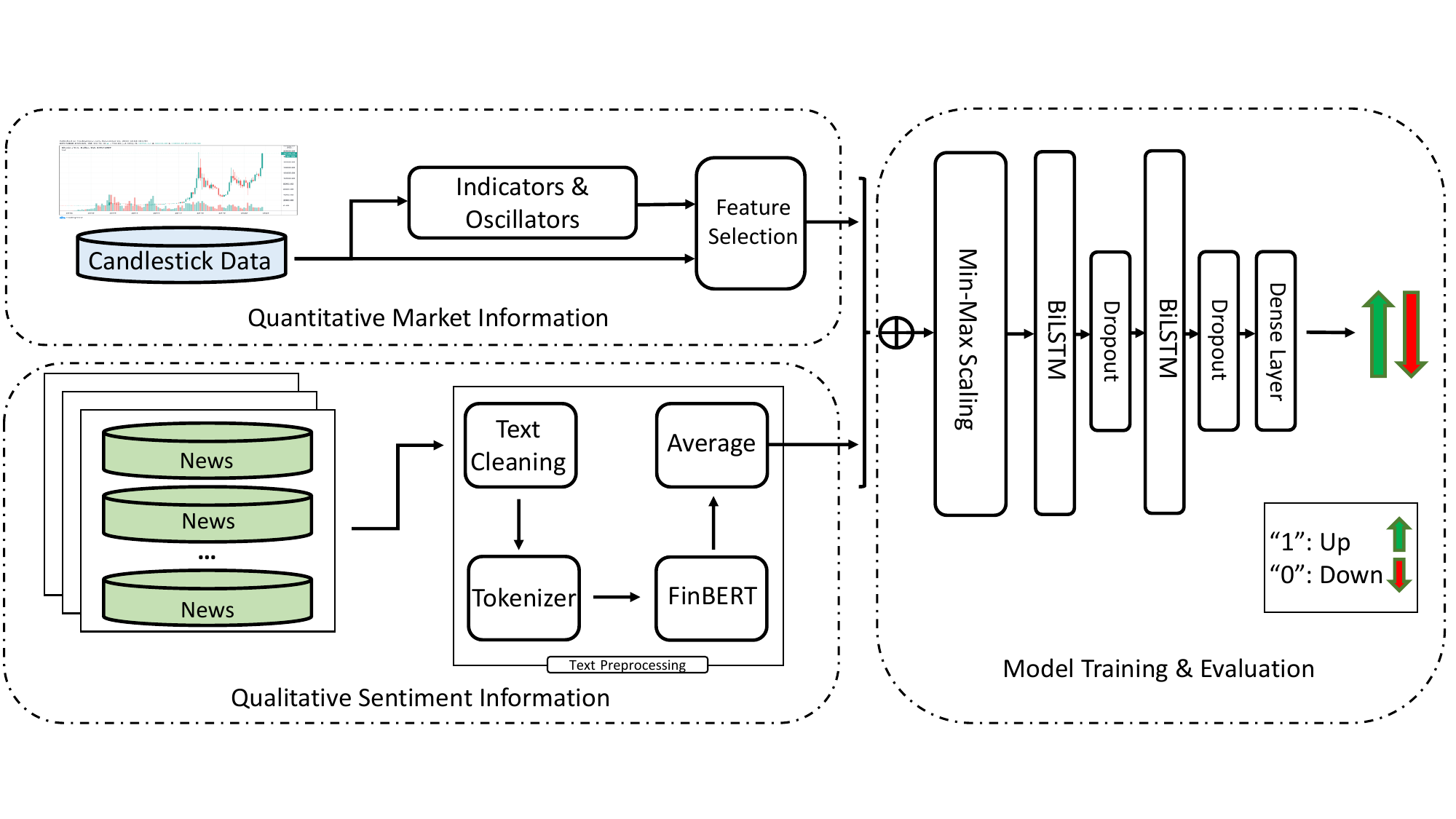}}
	\caption{Overview of the proposed framework.}
	\label{framework}
\end{figure*}

The proposed model consists of three components, as illustrated in Fig. \ref{framework}.

\subsection{Quantitative Market Information}
\label{Hard_Information}
Quantitative information refers to objective market data derived from historical trading activities and technical indicators. In the context of financial forecasting, particularly within cryptocurrency markets, such data provides the foundation for identifying patterns and informing predictive models.

Historical market data for Bitcoin were obtained from the publicly available Yahoo Finance platform\footnote{\url{https://finance.yahoo.com/}}, including daily open, high, low, and close prices as well as trading volumes. From this raw dataset, a comprehensive set of technical indicators was derived, widely recognized in literature for their effectiveness in financial time series forecasting \cite{kumbure2022machine, borges2020ensemble, ozer2022automated}.

An initial set of $53$ features, comprising trend-following, momentum, and volatility-based indicators, was constructed. To reduce redundancy and avoid multicollinearity, a correlation-based feature selection approach was applied. Specifically, the Pearson correlation coefficient $r_{i,j}$ between each pair of features $x_i$ and $x_j$ was calculated using:

\begin{equation}
    r_{i,j} = \frac{\sum_{k=1}^{n}(x_{i}^{(k)} - \bar{x}_i)(x_{j}^{(k)} - \bar{x}_j)}{\sqrt{\sum_{k=1}^{n}(x_{i}^{(k)} - \bar{x}_i)^2} \sqrt{\sum_{k=1}^{n}(x_{j}^{(k)} - \bar{x}_j)^2}}, \quad \text{for } i \neq j
    \label{eq:pearson}
\end{equation}

where $x_{i}^{(k)}$ and $x_{j}^{(k)}$ denote the values of features $i$ and $j$ on the $k$-th data instance, and $\bar{x}_i$, $\bar{x}_j$ represent their respective means. If $|r_{i,j}| > 0.95$, one of the two features was removed to minimize redundancy and improve model generalizability. This filtering process resulted in a final set of $36$ uncorrelated features.

Let $\mathbf{h}_{t} \in \mathbb{R}^{1 \times 36}$ denote the feature vector on trading day $t$, containing all selected technical indicators. To incorporate temporal dependencies, a rolling time window of size $T$ is constructed:

\begin{equation}
    \mathbf{H}_{t} = [\mathbf{h}_{t-T+1}, \dots, \mathbf{h}_{t-1}, \mathbf{h}_{t}], \label{Ht}
\end{equation}

where $\mathbf{H}_{t} \in \mathbb{R}^{T \times 36}$ represents the matrix of quantitative information used to predict the label $y_{t+1}$. Data points beyond the $T$-day window are excluded to maintain relevance and reduce the influence of outdated market dynamics.

\subsection{Qualitative Sentiment Information}
\label{Soft_Information}

Investor behavior and decision-making in financial markets are increasingly influenced by real-time information disseminated via social media platforms. Among these, X\footnote{\url{https://twitter.com/}} plays a significant role in reflecting market sentiment through news and user commentary related to cryptocurrencies \cite{zhou2023multi, boukhers2022ensemble}.

In this study, qualitative data refers to textual signals collected from X, specifically tweets related to Bitcoin. To ensure data quality and topical relevance, tweets were sourced from eight carefully selected accounts recognized for publishing timely and credible cryptocurrency news. The data were organized as a time series aligned with trading days.

Prior to sentiment analysis, standard preprocessing techniques were applied. User mentions (tokens containing ``$@$") were anonymized as ``$@user$", and hyperlinks (tokens containing ``$http$") were replaced with a generic placeholder ``$http$" to reduce noise and avoid bias in downstream processing.

RoBERTa \cite{liu2019roberta} and FinBERT \cite{araci2019finbert} are two pretrained transformer-based models utilized for sentiment extraction.
\begin{itemize}
    \item \textbf{RoBERTa} extends the BERT architecture with robust optimization and training on an expanded dataset, enabling deeper contextual understanding of text sequences.
    \item \textbf{FinBERT} is a domain-specific adaptation of BERT, fine-tuned on financial corpora such as earnings reports and financial news. This specialization enhances its performance in extracting sentiment within financial contexts.
\end{itemize}

Each tweet was tokenized and passed through the sentiment classifier, which outputs a probability distribution across three sentiment classes: positive, neutral, and negative. As multiple tweets are typically available for a single trading day, the corresponding sentiment scores were averaged to construct a daily sentiment vector, denoted by $s_t \in \mathbb{R}^{1 \times 3}$.

To capture temporal sentiment dynamics, qualitative information over a rolling window of $T$ trading days was aggregated into a matrix:

\begin{equation}
    \mathbf{S}_{t} = [s_{t-T+1}, \ldots, s_t] \in \mathbb{R}^{T \times 3}.
\end{equation}

This representation encodes sequential sentiment patterns and serves as a critical input to the fusion process with quantitative information for price movement prediction.

\subsection{Fusion of Quantitative and Qualitative Information for Model Evaluation}
\label{Hard_and_Soft_Information_Combination}

Following the extraction and preprocessing of both quantitative market indicators and qualitative sentiment signals, a unified dataset comprising numerical features is constructed. This dataset consists of two categories: one incorporating sentiment scores derived from FinBERT, and the other from RoBERTa. Accordingly, each feature vector resides in the space $\mathbb{R}^{39}$, and subsequent modeling procedures are conducted using this combined feature representation.

Let $T$ denote the rolling time window size. For each trading day $t$, the model input matrix $\mathbf{X}$ is formed by concatenating the corresponding quantitative information matrix $\mathbf{H}_t$ and qualitative information matrix $\mathbf{S}_t$ as follows:

\begin{equation}
    \mathbf{X} = [\mathbf{H}_t \oplus \mathbf{S}_t], \label{X}
\end{equation}

where $\oplus$ denotes the concatenation operation and $\mathbf{X} \in \mathbb{R}^{T \times 39}$. The sequence of concatenated daily feature vectors within $\mathbf{X}$ is represented as $\mathbf{x}_{t-T+1}, \ldots, \mathbf{x}_{t}$, with each vector $\mathbf{x}_t \in \mathbb{R}^{1 \times 39}$.

Prior to model training, the input data undergo Min-Max normalization to standardize the feature ranges. The normalized dataset is then partitioned into training, validation, and testing subsets using a stratified ratio of 70\%, 15\%, and 15\%, respectively. These subsets serve as input to the predictive models employed in the subsequent evaluation phase.

\subsection{Proposed Bitcoin Trend Prediction Model}
\label{Proposed_Bitcoin_Trend_Prediction_Model}

\begin{figure}[t]
    \centerline{\includegraphics[scale=0.5]{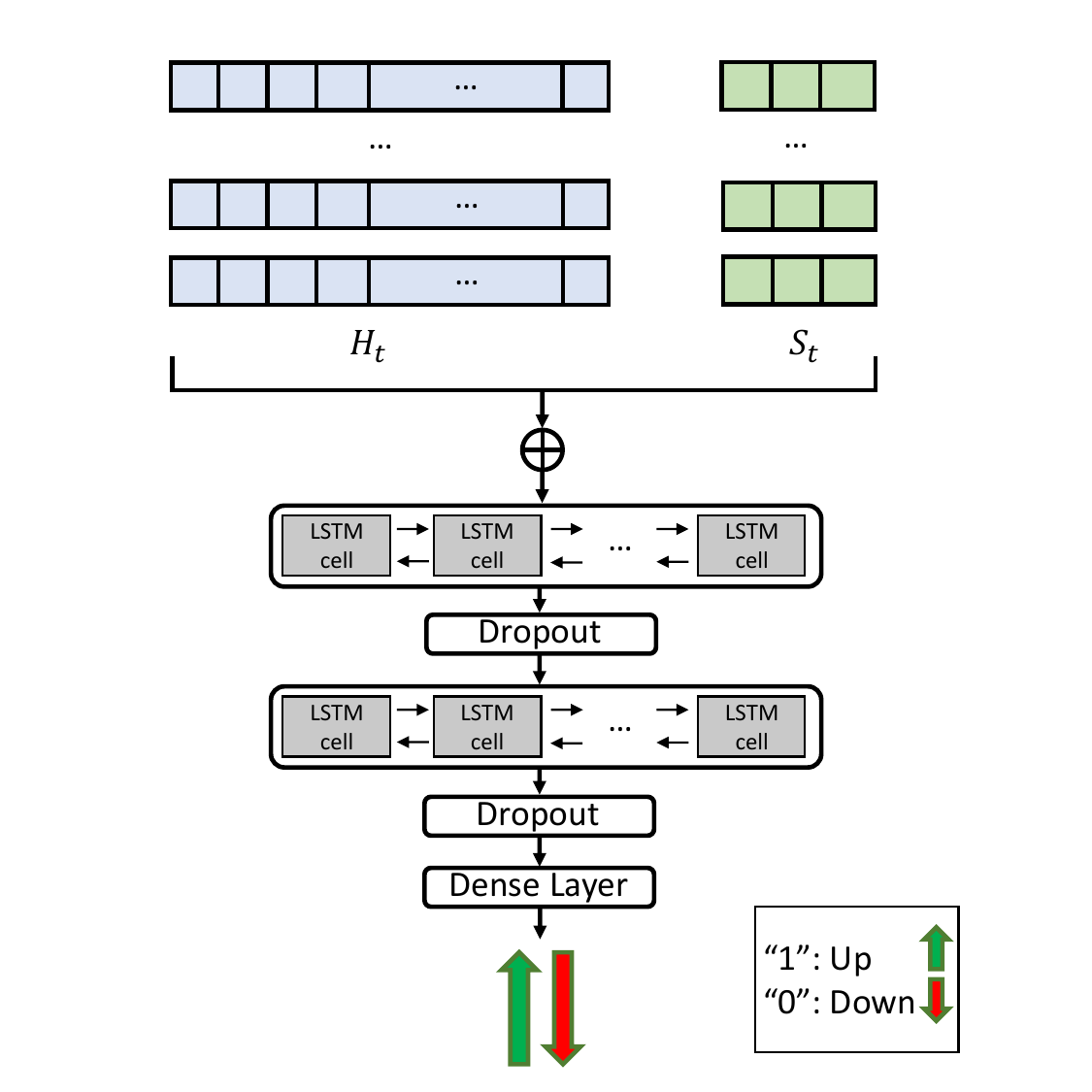}}
    \caption{The proposed model for final prediction of cryptocurrency price movements.}
    \label{BiLSTM}
\end{figure}

The proposed framework, illustrated in Fig. \ref{BiLSTM}, employs a two-layer BiLSTM architecture to capture sequential patterns relevant for cryptocurrency price movement prediction. BiLSTM networks are particularly effective in modeling long-term dependencies within time series, as they process input sequences in both forward and backward directions. This bidirectional mechanism enables the model to exploit the full temporal context, thereby identifying patterns that may be overlooked by unidirectional models.

The architecture can be formally described as:

\begin{equation}
    k_{t+1} = D(BiLSTM_{2}(D(BiLSTM_{1}(\mathbf{X})))), \label{Out_Put}
\end{equation}

where $BiLSTM_{1}$ and $BiLSTM_{2}$ represent the first and second BiLSTM layers, respectively, and $D(\cdot)$ denotes the dropout operation. The output $k_{t+1}$ comprises the concatenated forward and backward hidden states and is utilized to forecast the price movement on the subsequent trading day $t+1$.

To predict the probability of the next movement label $\hat{y}_{t+1}$, the vector $k_{t+1}$ is passed through a dense layer followed by a softmax activation function:
\begin{equation}
    \hat{y}_{t+1} = Softmax(Dense(k_{t+1})). \label{y_hat}
\end{equation}

The model is trained using the cross-entropy loss function:
\begin{equation}
    \mathcal{L} = CrossEntropy(\hat{y}_{t+1}, y_{t+1}), \label{Loss}
\end{equation}

where $y_{t+1} \in \{0, 1\}$ is the true binary label indicating whether the cryptocurrency price increases ($1$) or decreases ($0$) on day $t+1$.

\section{Experiments and Results}
\label{Experiments and Results}

\subsection{Datasets}

To evaluate the proposed model, Bitcoin, the most prominent and valuable cryptocurrency, was selected as the target asset. This selection is supported by the extensive availability of both quantitative market data and qualitative sentiment data relevant to Bitcoin over a long historical period.

Quantitative data were collected from the Yahoo Finance platform, encompassing daily open, high, low, and close prices as well as trading volumes. The final dataset spans from 6 April 2015 to 31 December 2022 and includes all relevant technical indicators derived from these raw inputs.

Qualitative data were sourced from the X platform. Initially, tweets from 12 publicly available accounts that regularly publish Bitcoin-related news were retrieved. After a filtering process to ensure topical relevance and reliability, data from 8 selected accounts were retained for analysis. Over the study period, a total of 104,296 news tweets related to Bitcoin were collected and processed.

The combined dataset was then chronologically partitioned into training (70\%), validation (15\%), and test (15\%) subsets. 

\subsection{Baselines}
To assess the effectiveness of the proposed model, a set of baseline approaches was implemented for comparative analysis. These include both classical ML algorithms and DL architectures, particularly those based on LSTM networks.

\begin{figure}[t]
    \centerline{\includegraphics[scale=0.6]{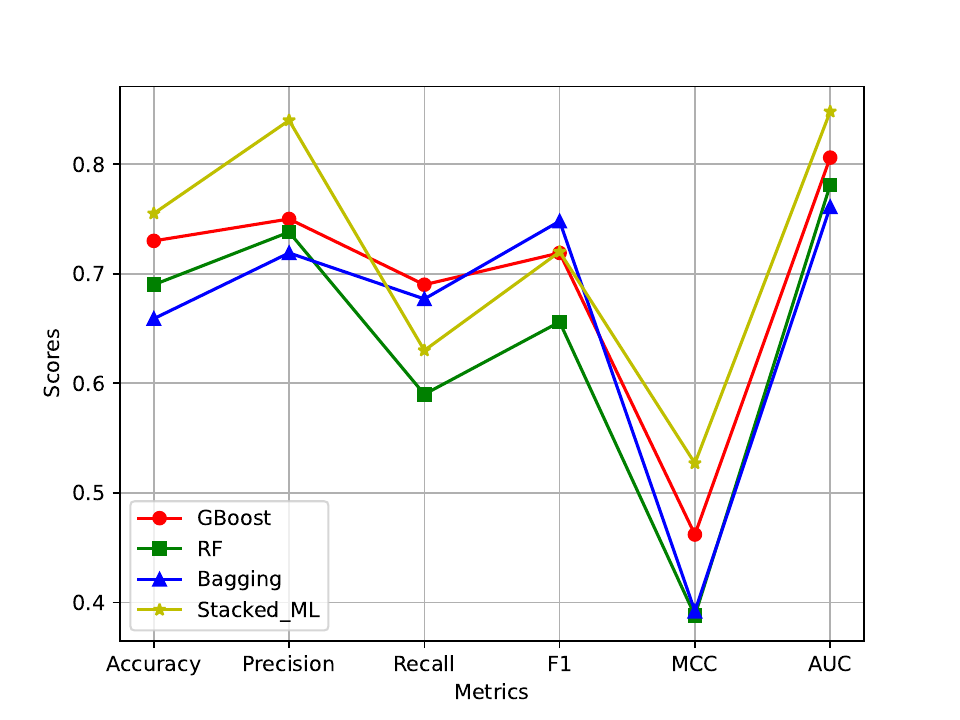}}
    \caption{Performance metrics for machine learning models.}
    \label{fig:ML}
\end{figure}

Preliminary experiments with individual ML classifiers yielded relatively weak performance compared to DL-based counterparts. To improve their effectiveness, an ensemble of three classifiers, Gradient Boosting (GBoost), Random Forest (RF), and Bagging, was constructed \cite{bara2024ensemble, chagahi2024cardiovascular, pourrezaee2024forecasting, dashtaki2025enhancing}. Nevertheless, even the ensemble model remained inferior to DL baselines. Consequently, only results derived from models utilizing fused quantitative and qualitative information (with sentiment signals processed by FinBERT) are reported. The corresponding evaluation metrics for ML models are illustrated in Fig. \ref{fig:ML}.

For a more comprehensive comparison, several LSTM-based DL models were selected as baselines, owing to their proven capacity to handle temporal dependencies and nonlinear patterns in time-series data. These models are described as follows:

\begin{itemize}
    \item \textbf{Stacked LSTM (StackedLSTM)}: This model extends the conventional LSTM architecture by stacking multiple LSTM layers. Each layer receives input from the preceding layer and forwards its output to the subsequent one, thereby enabling the model to learn hierarchical representations and capture more complex temporal dependencies \cite{ayitey2022forex}. In this study, the StackedLSTM model comprises three LSTM layers, each followed by a dropout layer with a dropout rate of $0.20$.

    \item \textbf{Convolutional Neural Network LSTM (CNNLSTM)}: The CNNLSTM model is a hybrid architecture that integrates convolutional layers for feature extraction and LSTM layers for modeling temporal sequences \cite{mao2024unveiling}. Convolutional layers identify local patterns in the input, while subsequent LSTM layers learn long-term dependencies. This model includes a CNN layer followed by three LSTM layers, each with a dropout layer of $0.20$.
    
    \item \textbf{Ordinary LSTM}: The basic LSTM model consists of sequentially arranged LSTM layers, each processing the input data and passing its output to the next layer \cite{koo2024centralized}. This architecture, while simpler than stacked or hybrid models, is effective for modeling time-dependent data. The implemented version includes two LSTM layers with dropout layers (rate of $0.20$) inserted after each LSTM layer.
\end{itemize}

\subsection{Setups}
\label{Setups}

The hyperparameters used in the experiments are summarized in Table \ref{Parameters}.

\begin{table}[H]
	\centering
	\caption{Experiment parameters.}
	\label{Parameters}
	\begin{tabular}{|l|c|}
        \hline
        \textbf{Parameter Name} & \textbf{Value} \\
        \hline
        Batch size & 16 \\
        \hline
        Maximum number of training epochs & 200 \\
        \hline
        Optimizer & Adam \\
        \hline
        Learning rate & 0.001 \\
        \hline
        Number of BiLSTM layers & 2 \\
        \hline
        Dimension of BiLSTM hidden states & 64 \\
        \hline
        Time step ($T$) & 21 \\
        \hline
        Dropout rate ($D(.)$) & 0.20 \\
        \hline
	\end{tabular}
\end{table}

To prevent overfitting, early stopping was applied based on validation loss. A grid search was performed to identify the optimal time step $T$, evaluated across $\{1, 5, 9, 13, 17, 21, 25, 29, 33\}$ and refined to $\{19, 20, 21, 22, 23\}$. In both cases, $T=21$ yielded the best results. Similarly, testing different BiLSTM depths showed that two layers offered the most effective balance between accuracy and model complexity. All reported results are based on this final configuration.

\subsection{Evaluation Metrics}
\label{Evaluation_Metrics}

To comprehensively evaluate model performance, both classification metrics and financial backtesting were considered. Classification performance was measured using accuracy, precision, recall, F1-score, and Matthews Correlation Coefficient (MCC). Their definitions are presented below.

\subsubsection{Accuracy}
Accuracy evaluates a model's prediction correctness by comparing the number of true positives and true negatives to the total number of predictions made.

\begin{equation}
\label{Accuracy}
Accuracy = \frac{TP+TN}{TP+TN+FP+FN}.
\end{equation}

\subsubsection{Precision}
In order to assess the accuracy of positive predictions, precision looks at the ratio of true positive predictions to all positive predictions made by a model.
\begin{equation}
\label{Precision}
Precision = \frac{TP}{TP+FP}.
\end{equation}

\subsubsection{Recall}
A model's recall rate, also called its sensitivity or true positive rate, measures the percentage of true positives correctly identified by the model. All positive occurrences can be captured by the model, according to the statement.   
\begin{equation}
\label{Recall}
Recall = \frac{TP}{TP+FN}.
\end{equation}

\subsubsection{F1-Score}
Taking into account the trade-off between recall and precision, the F1 score provides a well-balanced measure. Due to this quality, it is particularly useful for handling unbalanced datasets.
\begin{equation}
\label{F1}
F1-Score = \frac{2 \times TP}{2 \times TP + FP + FN}.
\end{equation}

\subsubsection{MCC}
    Unlike other metrics that tend to emphasize the positive class or are overly optimistic in imbalanced datasets, even if the classes are of different sizes, MCC provides a balanced measure.
\begin{equation}
\label{MCC}
\small
MCC = \frac{TP \times TN - FP \times FN}{\sqrt{(TP + FP)(TP +FN)(TN + FP)(TN + FN)}},
\end{equation}

where $TP$ is true positive, $FP$ is false positive, $TN$ is true negative, and $FN$ is false negative. 

For financial evaluation, a backtesting simulation is conducted based on the predicted cryptocurrency price movements. This simulation aims to assess the practical utility of the model in a real-world trading scenario.

\subsubsection{Total Money (TMoney)}
The TMoney metric is used to evaluate the financial performance of the trading strategy simulated through backtesting \cite{ma2023multi, ma2024quantitative}. An initial capital of 100,000 USD was assumed, with a 0.1\% transaction commission. A Buy \& Hold strategy was used as a benchmark.  

The trading rules were:
\begin{itemize}
    \item If the predicted label for day $t+1$ was upward (1) and differed from the previous day, all capital was invested at the close of day $t$. 
    \item If the prediction was downward (0) and differed from the previous day, all holdings were sold. 
    \item Otherwise, no trade was executed. 
\end{itemize}

This decision logic ensures that trading actions are taken only when a change in trend is detected, reducing unnecessary transaction costs.

\begin{figure}[H]
    \centerline{\includegraphics[scale=0.5]{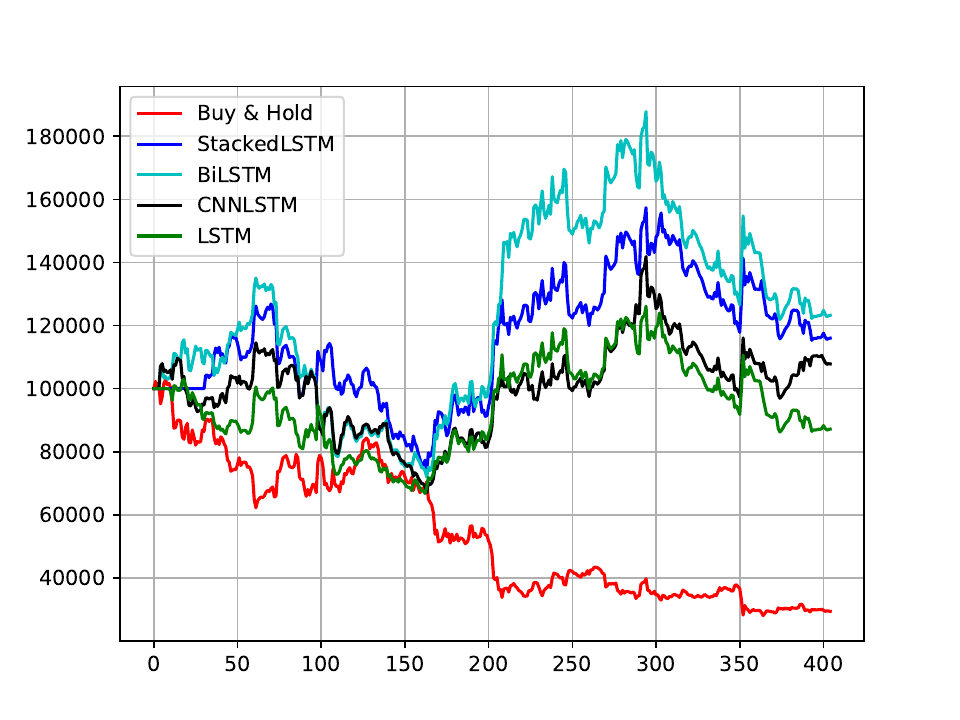}}
	\caption{Cumulative trading capital (TMoney) of different models.}
	\label{TMoney}
\end{figure}

Fig. \ref{TMoney} shows cumulative TMoney over time, comparing different models. The proposed BiLSTM consistently outperforms all baselines and Buy \& Hold.

\begin{figure*}[t]
    \centering
    \begin{minipage}{0.45\textwidth}
        \centering
        \subfloat[LSTM\label{subfig:LSTM}]{
            \includegraphics[width=\textwidth]{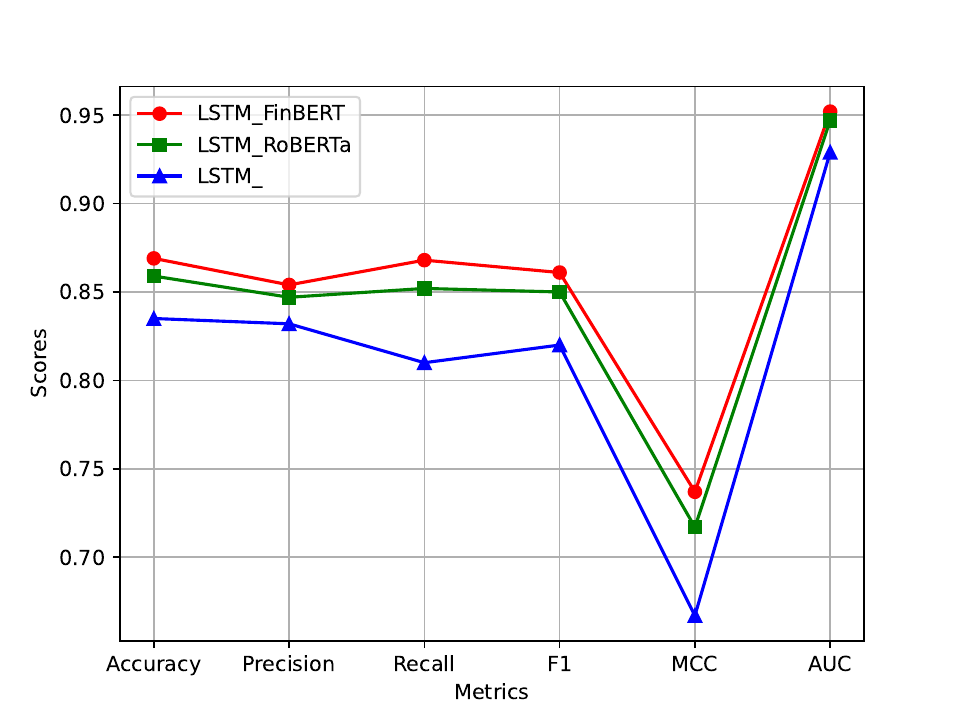}
        }
    \end{minipage}
    \hfill
    \begin{minipage}{0.45\textwidth}
        \centering
        \subfloat[BiLSTM (Proposed model)\label{subfig:BiLSTM}]{
            \includegraphics[width=\textwidth]{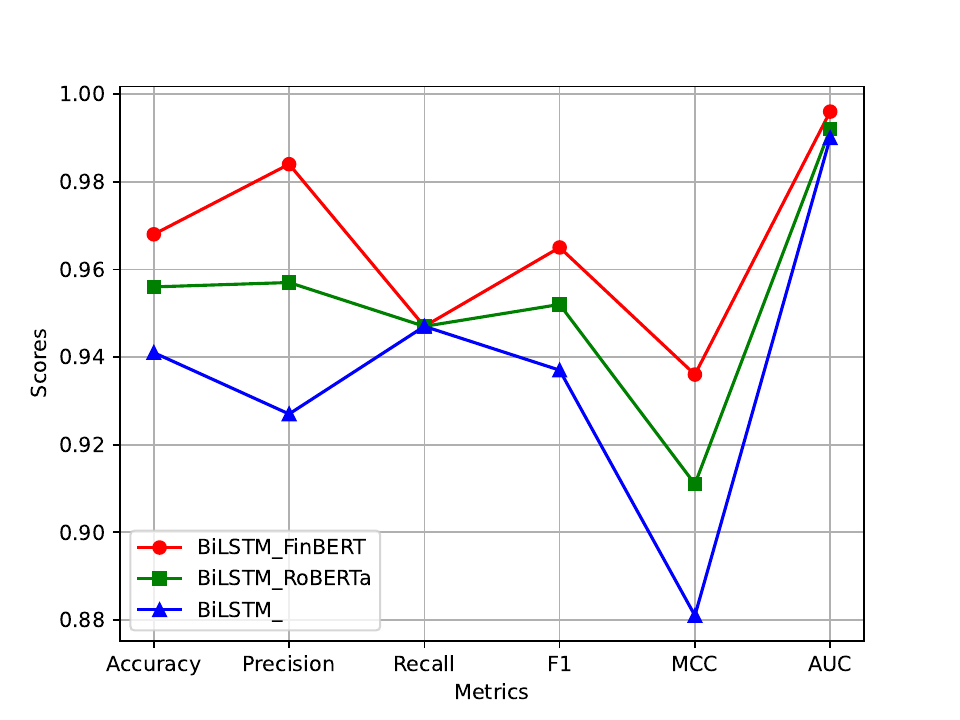}
        }
    \end{minipage}

    \vspace{0.5cm}

    \begin{minipage}{0.45\textwidth}
        \centering
        \subfloat[CNNLSTM\label{subfig:CNNLSTM}]{
            \includegraphics[width=\textwidth]{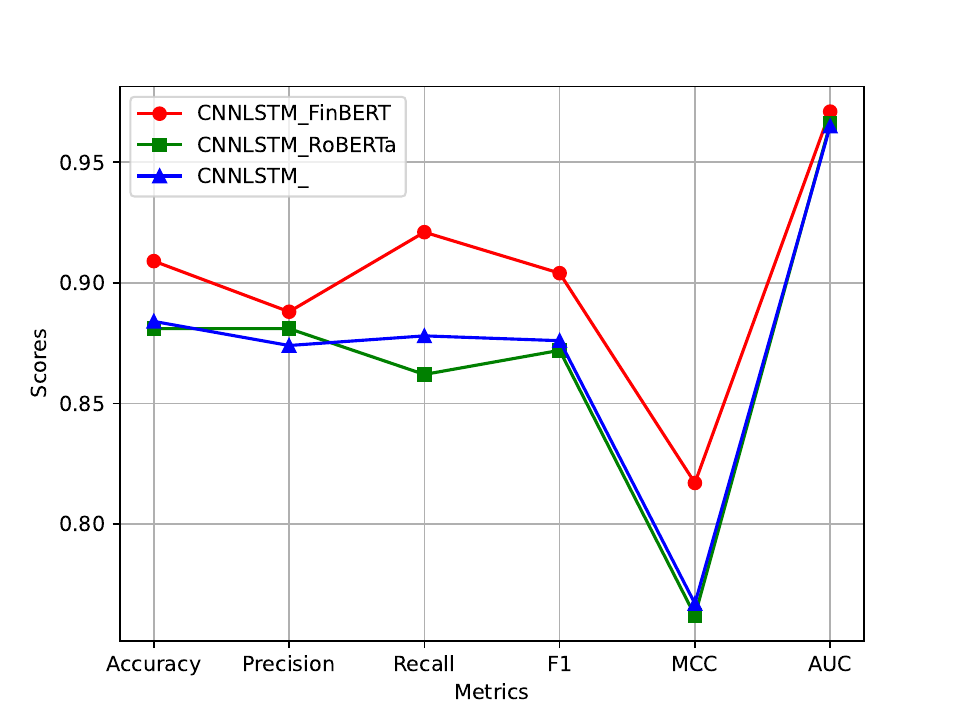}
        }
    \end{minipage}
    \hfill
    \begin{minipage}{0.45\textwidth}
        \centering
        \subfloat[StackedLSTM\label{subfig:Stacked_LSTM}]{
            \includegraphics[width=\textwidth]{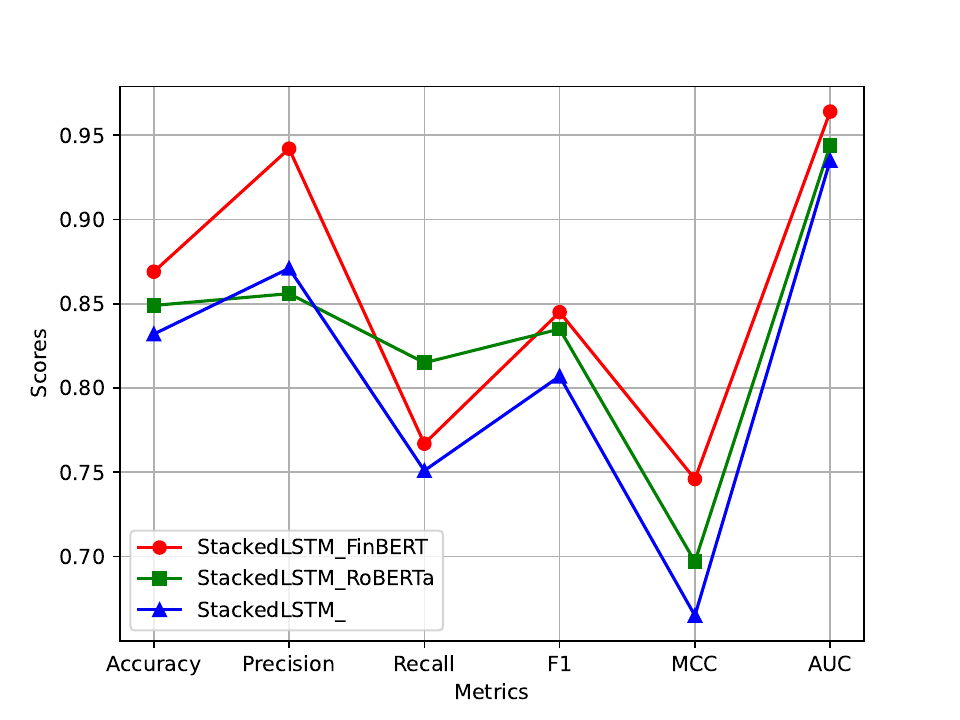}
        }
    \end{minipage}

    \caption{Performance metrics results of models: (a) LSTM, (b) BiLSTM (Proposed model), (c) CNNLSTM, and (d) StackedLSTM.}
    \label{fig:Results}
\end{figure*}

Fig. \ref{fig:Results} reports classification metrics across DL models, showing that the proposed BiLSTM achieves the highest and most consistent performance.






Overall, the superior performance of the proposed BiLSTM model can be attributed to its ability to capture sequential dependencies in financial time series, incorporate market sentiment, and effectively fuse quantitative and qualitative information. As shown in Figs. \ref{fig:Results} and \ref{TMoney}, the proposed model consistently outperforms alternatives across both predictive accuracy and financial profitability.

Two key observations can be drawn from these results. First, integrating quantitative indicators with sentiment signals derived from social media yields substantially better performance than models relying exclusively on market-based features. This finding underscores the added predictive value of incorporating real-time financial sentiment into forecasting frameworks. Second, models leveraging sentiment representations extracted with FinBERT consistently achieve higher accuracy and robustness than those based on RoBERTa, highlighting the advantage of domain-specific language models in financial applications.

Furthermore, Fig. \ref{TMoney} illustrates that the proposed BiLSTM model surpasses all baseline methods as well as the Buy \& Hold strategy in terms of cumulative financial return (TMoney). The backtesting results confirm that combining BiLSTM with FinBERT-based sentiment signals leads to more profitable trading decisions and superior investment performance over time.

\section{Future Research Directions}
\label{Future Research Directions}

Although the proposed model demonstrates strong performance in predicting cryptocurrency price movements, several avenues remain open for future exploration to further enhance robustness, applicability, and interpretability.

\begin{itemize}
    \item \textbf{News Impact Weighting:} In the current framework, all news items are treated equally during sentiment aggregation. Future studies could develop importance-weighting mechanisms to dynamically assign relevance scores to news articles based on their credibility, market impact, or source authority.

    \item \textbf{Multilingual Sentiment Analysis:} The present study is limited to tweets and headlines in English. Extending the sentiment analysis pipeline to support multilingual content would improve coverage and predictive power, particularly in global cryptocurrency markets.
\end{itemize}

\section{Conclusion}
\label{Conclusion}
Accurate prediction of cryptocurrency price movements remains a challenging task due to the volatile and multifactorial nature of digital asset markets. This study introduced a novel multisource fusion framework that integrates quantitative financial indicators with sentiment signals extracted from social media to improve the accuracy and robustness of cryptocurrency trend prediction. 

The proposed architecture combines technical indicators with sentiment features derived from a FinBERT-based analysis of news and tweets, which are jointly modeled using a BiLSTM network to capture bidirectional temporal dependencies. Bitcoin, as the most prominent and data-rich cryptocurrency, was employed as the experimental case study. Empirical results demonstrated that the proposed framework consistently outperformed existing baselines, achieving improvements of at least 5.9\% in accuracy and 11.9\% in MCC. Beyond classification performance, the model also yielded superior financial returns in backtesting compared to traditional DL models and the Buy \& Hold strategy. 

These findings highlight the importance of integrating heterogeneous information sources, particularly real-time market sentiment, with traditional indicators to enhance predictive performance. The proposed approach offers practical value for both investors and algorithmic trading systems by enabling more informed decision-making and reducing risks associated with highly volatile cryptocurrency markets.

\bibliography{references}

\end{document}